\documentclass{article}
\usepackage{graphicx}

\usepackage{arxiv}

\usepackage[utf8]{inputenc} % allow utf-8 input
\usepackage[T1]{fontenc}    % use 8-bit T1 fonts
\usepackage{hyperref}       % hyperlinks
\usepackage{url}            % simple URL typesetting
\usepackage{booktabs}       % professional-quality tables
\usepackage{amsfonts}       % blackboard math symbols
\usepackage{nicefrac}       % compact symbols for 1/2, etc.
\usepackage{microtype}      % microtypography
\usepackage{lipsum}		% Can be removed after putting your text content
\usepackage{graphicx}
\usepackage{doi}

\usepackage[
backend=biber,
style=chem-acs,
]{biblatex}
\title{BASIL: Bayesian Application for Scientific Iteration and Learning}

\date{} 	
\author{
Kelvin P. Idanwekhai$^{1,2}$ \hspace{0.5em} Valeriia Kaneva$^{2}$ \hspace{0.5em} Stefano Menegatti$^{3,4}$ \quad \textbf{Alexander Tropsha}$^{2}$  \\
$^1$ Department of Chemistry, UNC Chapel Hill, Chapel Hill, NC, USA \\
$^2$ Laboratory for Molecular Modeling (MML), UNC Chapel Hill, Chapel Hill, NC, USA\\ 
$^3$ Department of Chemical and Biomolecular Engineering, NC State University, Raleigh, NC, USA \\
$^4$ ChromaGenix, Raleigh, NC, USA \\
\texttt{\{kelidan,alex\_tropsha\}@unc.edu}\\
}

\usepackage{amsmath}
\begin{document}
\maketitle

\date{} 					% Or removing it

% \author{ \href{https://orcid.org/0000-0002-3804-8830}{\includegraphics[scale=0.06]{orcid.pdf}\hspace{1mm}Kelvin P. Idanwekhai} \\
% 	Department of Chemistry\\
% 	 University of North Carolina at Chapel Hill\\
% 	Chapel Hill, NC \\
% 	\texttt{kelidan@unc.edu} \\
% 	%% examples of more authors
% 	\And
% 	\href{https://orcid.org/0000-0002-1594-8182}{\includegraphics[scale=0.06]{orcid.pdf}\hspace{1mm}Valeriia Kaneva} \\
%   Division of Chemical Biology and Medicinal Chemistry,\\
% 	 UNC Eshelman School of Pharmacy,\\
% 	Chapel Hill, NC \\
% 	\texttt{valk@email.unc.edu} \\
%     \And
% 	\href{https://orcid.org/0000-0001-5633-434X}{\includegraphics[scale=0.06]{orcid.pdf}\hspace{1mm}Stefano Menegatti} \\
% 	Department of Chemical and Biomolecular Engineering\\
% 	 NC State University\\
% 	Raleigh, NC \\
% 	\texttt{smenega@ncsu.edu} \\
%     \And
% 	\href{https://orcid.org/0000-0003-3802-8896}{\includegraphics[scale=0.06]{orcid.pdf}\hspace{1mm}Alexander Tropsha} \\
% 	Division of Chemical Biology and Medicinal Chemistry,\\
% 	 UNC Eshelman School of Pharmacy,\\
% 	Chapel Hill, NC \\
% 	\texttt{alex_tropsha@unc.edu} \\
% }

% Uncomment to remove the date
% \date{}

% Uncomment to override  the `A preprint' in the header
\renewcommand{\headeright}{Preprint}
% \renewcommand{\undertitle}{Preprint}
% \renewcommand{\shorttitle}{\textit{arXiv} Template}

% \hypersetup{
% pdftitle={BASIL: Bayesian Application to Scientific Iteration and Learning},
% pdfauthor={Kelvin P. Idanwekhai, Valeriia Kaneva, Stefano Menegatti, and Alexander Tropsha},
% pdfkeywords={First keyword, Second keyword, More},
% }

% \begin{document}
%%\maketitle

\begin{abstract}
We introduce BASIL, a user-friendly desktop application for process optimization. BASIL employs a Bayesian approach, incorporating special acquisition functions that can be used to solve both single and multi-objective optimization problems. It provides a graphical interface that enables users to input their experimental parameters, optimization objectives, and legacy data. This is then used to build surrogate models, which are coupled with acquisition functions to guide and optimize a process towards a desired objective. To facilitate model building, BASIL provides a variety of predefined surrogate model templates. BASIL can be used to optimize any arbitrary experiment or process with known, user-defined input variables, optimization objectives, and defined output.
\end{abstract}

% keywords can be removed
% \keywords{Bayesian optimization, \and Process modelling \and Surrogate model}

\section{Introduction}
Many processes, ranging from small-scale scientific experiments to
large-scale industrial production, require the optimization of process
parameters. Historically, process optimization has been driven by a
combination of numerical, analytical, and empirical methods. Commonly
used techniques include factorial design, simulated annealing, and
response surface
methodology\cite{greenhillBayesianOptimizationAdaptive2020}. While these
methods have shown success in specific fields, their inability to
generalize and provide interpretable results continues to be problematic
\cite{greenhillBayesianOptimizationAdaptive2020}.

Recently, machine learning (ML) methods have become increasingly used
for process optimization. Among those, Bayesian optimization (BO), which
combines model-based function approximators with heuristic acquisition
functions, has been especially successful as applied to process
optimization\cite{bennettAutonomousReactionParetofront2024,greenhillBayesianOptimizationAdaptive2020,macleodSelfdrivingLaboratoryAdvances2022}.
By balancing the exploration of promising experimental design regions
with the exploitation of historic data, BO algorithms can adapt and
converge toward optimal process conditions, minimizing the number of
experiments required and maximizing both the objective and efficiency of
the process\cite{frazierTutorialBayesianOptimization2018}. One of the
most valuable features of this approach is its ability to generalize
without any mechanistic assumptions about the process, and more
importantly, show high optimization efficiency using a relatively small
dataset\cite{frazierTutorialBayesianOptimization2018}.

Machine learning approaches used for process optimization commonly
employ algorithms such as Gaussian processes (GPs) and quasi-Monte Carlo
acquisition
functions\cite{luSurrogateModelingBayesian2022,rochChemOSOrchestrationSoftware2018,rooneySelfdrivingLaboratoryDesigned2022,shieldsBayesianReactionOptimization2021}.
The need for these methods stems from their ability to efficiently
navigate noisy, non-convex landscapes with limited samples, making them
particularly valuable for both scientific optimization and discovery,
where evaluating process parameters experimentally is expensive.

The BO framework has been applied to domains such as regenerative
medicine\cite{kandaRoboticSearchOptimal2022}, materials
discovery\cite{abolhasaniRiseSelfdrivingLabs2023}, processing of
biological therapies\cite{idanwekhaiAdaptiveMachineLearning2026}, and
biosystems design\cite{hamediradFullyAutomatedAlgorithm2019}. While BO
is a useful method, successfully implementing and integrating it into an
experimental workflow can be challenging for scientists without
knowledge of programming and ML. Therefore, providing an easy interface
to these methods is critical to enabling wider adoption by experimental researchers.

\begin{figure}
	\centering
	\includegraphics[width=1\linewidth]{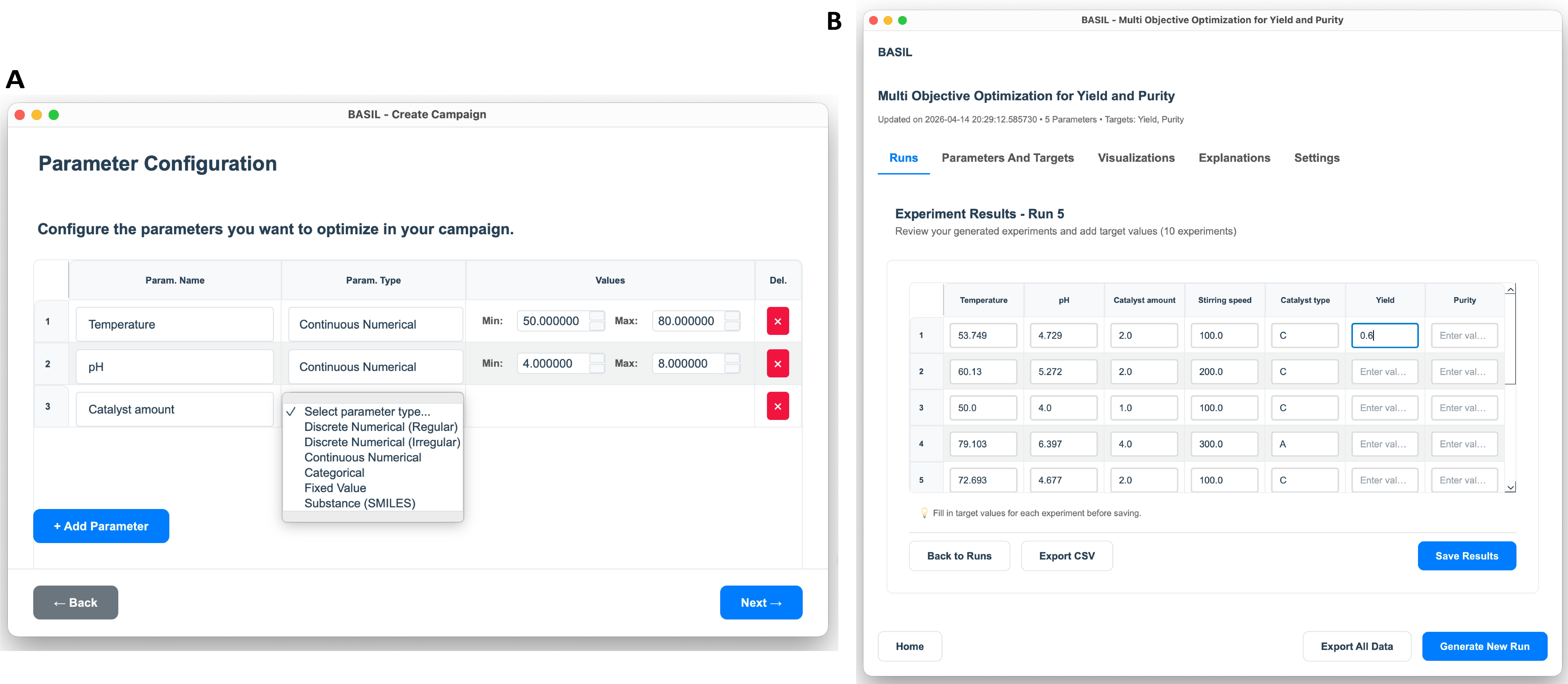}
    \label{fig:fig1}
	\caption{Campaign creation and management interface. (A) Campaign configuration panel, in which the user defines campaign parameters by specifying name, type, and search space boundaries. The screenshot shows two parameters already defined: Temperature — a continuous parameter greater than or equal to 50 and less than 80; pH — a continuous parameter greater than or equal to 4 and less than 8. The third parameter, Catalyst amount, is defined, but its type and possible values have not yet been specified. (B) Experimental data entry interface, displaying model-recommended parameter combinations and fields for outcome entry. The screenshot shows the parameter combinations for run 5, with specified values for Temperature, pH, Catalyst amount, Stirring speed, and Catalyst type. Values can be manually modified or exported as a CSV file (Export CSV button). The interface allows users to enter, save, and modify experimental results at any stage before finalizing a run. The Generate New Run button initiates the next optimization iteration based on the entered results.}
\end{figure}

There are several libraries and frameworks tailor-made for BO and
model-based optimization:
ProcessOptimizer\cite{bertelsenProcessOptimizerOpenSourcePython2025},
Bayesian Backend (BayBE)\cite{fitznerBayBEBayesianBack2025a},
BOTorch\cite{balandatBoTorchFrameworkEfficient2020},
BOFire\cite{durholtBoFireBayesianOptimization2025}, etc. While
efficient, these tools require the user to have a working knowledge of
either Python, the command line, or ML algorithms. This creates an
adoption hurdle for scientists since they would either need a dedicated
data science team or spend significant time learning how to use these
methods. ProcessOptimizer, which positions itself as an easier
alternative due to its simple Python API, lacks integrated acquisition
functions and does not support categorical variables in the parameter
space\cite{bertelsenProcessOptimizerOpenSourcePython2025}. BASIL solves
this problem by offering model-based optimization strategies in an
intuitive interface (Fig. \ref{fig:fig1}). It includes surrogate models and acquisition
functions for both single and multi-objective optimization problems. We
use BayBE\cite{fitznerBayBEBayesianBack2025a} to support the BO routines,
as it offers an interoperable API. BASIL currently supports Windows,
Mac, and Linux operating systems

\section{Design and Implementation}

\subsection{Software Design}

BASIL implements several specific considerations concerning its design
and functionality as follows:

\begin{itemize}
\item
  Unlike other optimization programs with a steep learning curve, BASIL
  is user friendly. The primary interface is simplified to enable users
  to merely enter their input data and launch calculations.
\item
  This intuitive interface makes BASIL a general framework, applicable
  to a wide variety of experimental or process optimization problems.
  Once the users define the inputs and outputs for their process of
  interest, they can launch the software.
\item
  The topic of proper research data management has become popular in the
  last decade, especially with the advances of artificial intelligence
  and curation of multi-fidelity data. BASIL manages and keeps track of
  all experimental data generated by the software or uploaded to it.
  This helps make downstream reporting and analysis easier.
\item
   We adhere to the ``file over apps'' philosophy\cite{angoFileApp2023}.
  BASIL does not use or generate any proprietary data formats; instead,
  all data and results are stored in JSON and CSV formats. This means
  that users can transfer data files for use in other programs, while
  maintaining data compatibility.
\end{itemize}

\subsection{Features of BASIL}

\begin{enumerate}
\def\labelenumi{\alph{enumi})}
\item
  It offers an intuitive interface for scientists and technicians to
  perform model-based optimization without writing any code. We also
  provide stable defaults that can be used for most processes, reducing
  the decision fatigue when trying to quickly optimize a process.
\item
  It incorporates multiple surrogate models and acquisition functions
  that can be selected by skilled users. This makes it easy to benchmark
  on a variety of different modeling and sampling methods.
\item
  Users can solve both single- and multi-objective optimization
  problems. In the case of multi-objective optimization, it is possible
  to opt for optimizing the objectives as a weighted sum or finding the
  Pareto frontier of all objectives.
\item
  The interface supports process parameter visualizations and model
  explanation. This functionality is useful for report generation and
  understanding of model-guided process optimization.
\end{enumerate}

\subsection{Implementation}

\begin{enumerate}
\def\labelenumi{\alph{enumi})}
\item
  Surrogate model: A GP surrogate is implemented along with other
  ensemble models. In the case of GPs, we provide support for pluggable
  kernel choices and use numerically robust matrix routines to compute
  posterior mean and variance for candidate inputs.
\item
  Acquisition and optimization: Acquisition functions (e.g., Expected
  Improvement, Upper Confidence Bound) are implemented alongside numeric
  optimizers. The acquisition optimizer uses a mixture of random
  restarts and local gradient-based optimization to locate high-value
  candidates under box constraints; discrete and categorical parameters
  are handled via enumerations and one-hot encodings before
  optimization.
\item
  Parameter handling and transformations: Input parameter spaces, value
  normalization, and encoding/decoding routines are provided. Continuous
  parameters are normalized to unit ranges for GP input; categorical
  parameters use ordinal or one-hot encodings, depending on downstream
  kernel compatibility. The system preserves original parameter metadata
  (bounds, types, priors) so proposed candidates can be round-tripped to
  experiment controllers.
\item
  Objective evaluation \& orchestration: It provides an interface to run
  external experiments, ingests results, and returns objective values.
  The architecture separates synchronous local evaluations from
  asynchronous or remote experiment runners, so the same BO core can be
  used with simulated test problems or real lab hardware.
\end{enumerate}

\section{Results}

To demonstrate the functionality of the developed software, we present a
synthetic case study of multi-objective chemical reaction optimization.
The process includes five parameters: two continuous (temperature, pH),
two discrete (catalyst amount, stirring speed), and one categorical
(catalyst type) (Table \ref{tab:table1}). The two objectives are optimized yield and
purity. To simulate a real experiment, we compute the objective
functions as weighted sums of quadratic functions for the numerical
parameters and constant values for the categorical parameter (Eq. \ref{eq:equation1};
Tables \ref{tab:table1},\ref{tab:table2},\ref{tab:table3}).

\begin{equation}
Yield(T,\ p,\ c,\ s,\ t) = 0.30 \cdot S_{T}^{Y}\  + \ 0.25 \cdot S_{p}^{Y} + 0.20 \cdot S_{c}^{Y} + \ 0.15 \cdot S_{S}^{Y} + \ 0.10 \cdot S_{t}^{Y} \\
\label{eq:equation1}
\end{equation}

\begin{equation}
Purity(T,\ p,\ c,\ s,\ t) = 0.35 \cdot S_{T}^{P}\  + \ 0.25 \cdot S_{p}^{P} + 0.15 \cdot S_{c}^{P} + \ 0.15 \cdot S_{S}^{P} + \ 0.10 \cdot S_{t}^{P}
\label{eq:equation2}
\end{equation}

\begin{equation}
{S_{x}^{i} = \ S}_{i}(x) = \  - a_{i}\left( x - x_{opt} \right)^{2} + 100\ 
\label{eq:equation3}
\end{equation}

\textbf{Objective functions defining Yield and Purity
(first and second equations, respectively).} \(T\), \(p\), \(c\), \(s\),
\(t\) denote the parameters described in (Table \ref{tab:table1}). All numerical
parameters contribute via quadratic component functions \(S_{x}^{i}\)
(third equation; parameters listed in Table \ref{tab:table2}). Catalyst type scores are
defined as constants (Table \ref{tab:table3}).

\begin{table}
	\caption{Optimization campaign parameter space (arbitrary units).}
	\centering
	\begin{tabular}{lll}
		\toprule
		% \cmidrule(r){1-2}
		Parameter     & Symbol     & Range \\
		\midrule
		Temperature & \(T\) & \(\lbrack 50,\ 90\rbrack\) \\
pH & \(p\) & \(\lbrack 4,\ 8\rbrack\) \\
Catalyst amount & \(c\) & \(\{ 1,\ 2,\ 3,\ 4,\ 5\}\) \\
Stirring speed & \(s\) & \(\{ 100,\ 200,\ 300,\ 400\}\) \\
Catalyst type & \(t\) & \(\{ A,\ B,\ C\}\) \\
		\bottomrule
	\end{tabular}
	\label{tab:table1}
\end{table}

\begin{table}
	\caption{Parameters of the synthetic objective functions (arbitrary units).}
	\centering
	\begin{tabular}{lllll}
		\toprule
		% \cmidrule(r){1-2}
		Parameter & Yield optimum \(\mathbf{x}_{\mathbf{i}}\) & Purity optimum \(\mathbf{x}_{\mathbf{i}}\) & Yield scale coefficient \(\mathbf{a}_{\mathbf{i}}\) & Purity scale coefficient \(\mathbf{a}_{\mathbf{i}}\) \\
		\midrule
Temperature & 80 & 60 & 0.025 & 0.030 \\
pH & 6.5 & 5.5 & 6.5 & 5.5 \\
Catalyst amount & 4 & 3 & 10 & 10 \\
Stirring speed & 300 & 200 & 0.0015 & 0.0020 \\
		\bottomrule
	\end{tabular}
	\label{tab:table2}
\end{table}

\begin{table}
	\caption{Catalyst type contribution scores for Yield and Purity
objective functions (arbitrary units).}
	\centering
	\begin{tabular}{lll}
		\toprule
		% \cmidrule(r){1-2}
		Catalyst Type & Yield \(\mathbf{S}_{\mathbf{t}}^{\mathbf{Y}}\) & Purity \(\mathbf{S}_{\mathbf{t}}^{\mathbf{P}}\) \\
		\midrule
A & 100 & 70 \\
B & 70 & 100 \\
C & 85 & 85 \\
		\bottomrule
	\end{tabular}
	\label{tab:table3}
\end{table}

\begin{figure}
	\centering
    \label{fig:fig2}
	\includegraphics[width=1\linewidth]{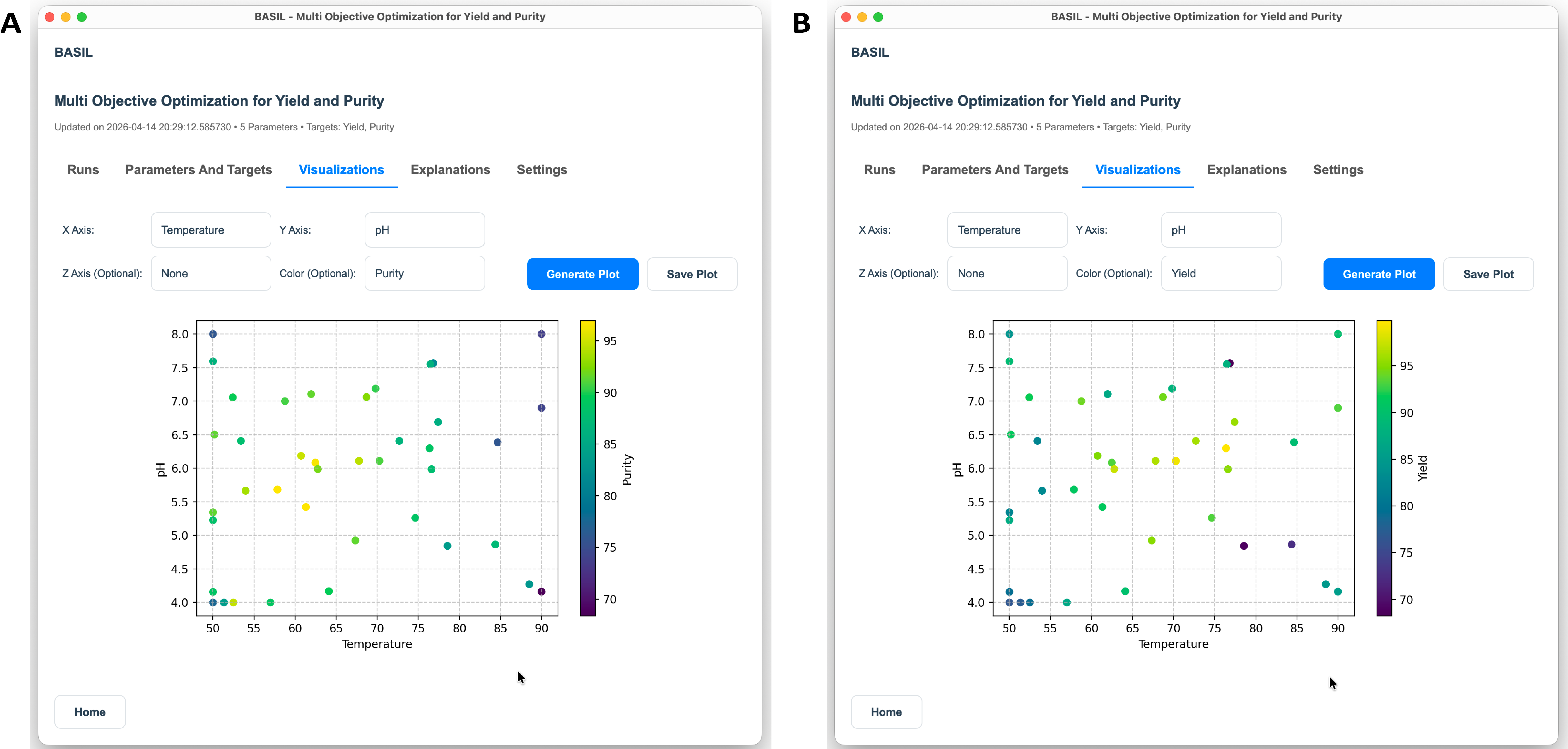}
	\caption{ Campaign results visualization interface. The Visualization tab allows visualization of relationships between experimental parameters and outcomes in 2D and 3D, including colormaps. In (A), the X axis represents Temperature, the Y axis represents pH, and color indicates Purity for the corresponding parameter values. In (B) the color reflects Yield. The interface also allows specifying a Z axis for 3D visualization, and plots can be exported as files.}
\end{figure}

First, in the GUI, the user specifies the optimization strategy
(multi-objective, Pareto), the surrogate model type (Gaussian Process
with Radial Basis function by default), and the acquisition function,
such as \emph{logNoisyExpectedHypervolumeImprovement} (logNEHVI).
Parameters, their types, and boundaries are then specified (Fig. \ref{fig:fig1}A).
Optionally, existing experimental data can be imported. The
configuration appears in the users' workspace as a named campaign and is
ready for initialization.

If no preexisting experimental data is available, a random set of
experimental parameters are recommended as the first batch for initial
space exploration. These experiments are accessible in the interface and
can be downloaded as a CSV file. After conducting experiments, the
results are entered into the interface (Fig. \ref{fig:fig1}B). Once all results are
entered, a new batch of experiments can be generated. New parameter
values are suggested by the acquisition function based on the previous
experimental results to improve the objectives.

\begin{figure}
	\centering
    \label{fig:fig3}
	\includegraphics[width=1\linewidth]{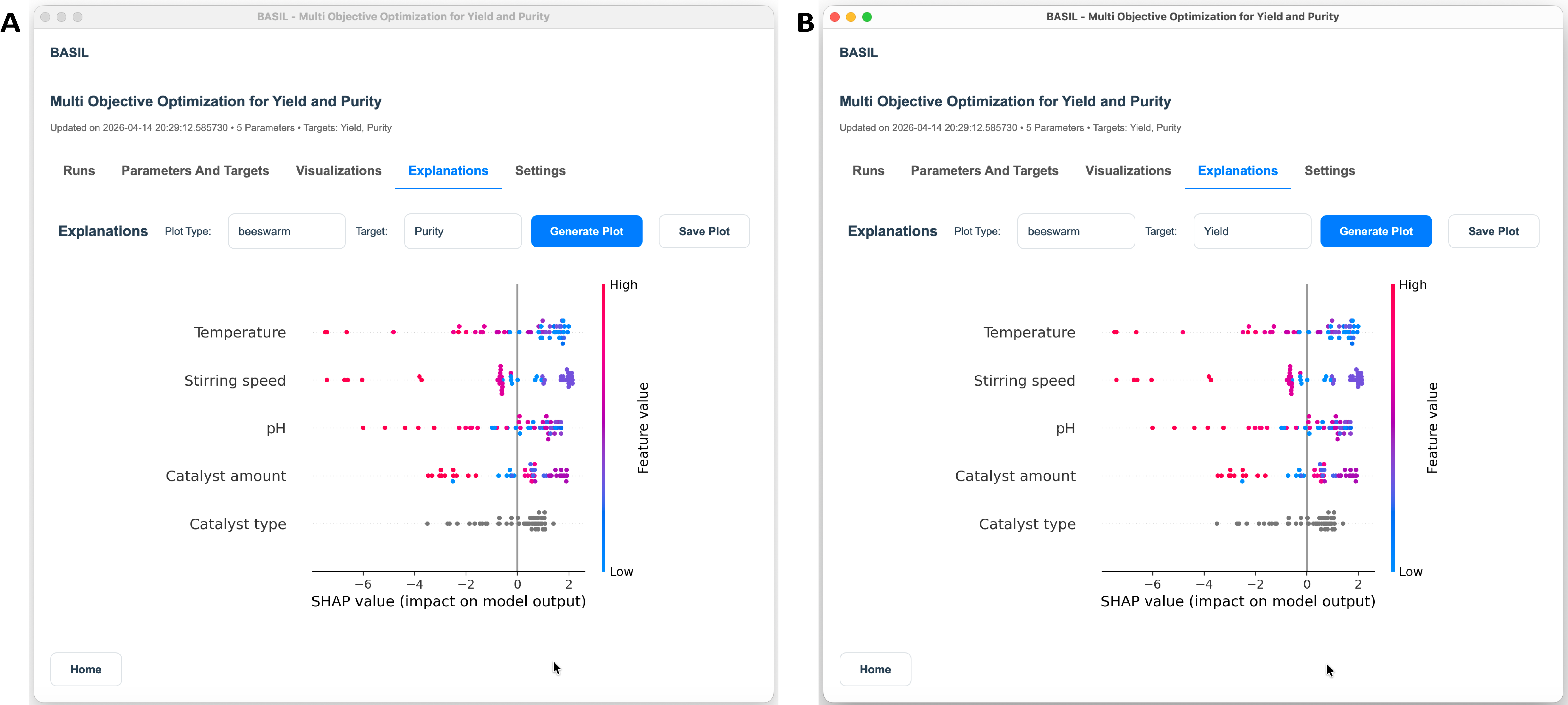}
	\caption{Campaign results explanation interface. The Explanations tab provides functionality to visualize SHAP (Shapley Additive Explanation) values to quantify how each parameter contributes to the outcome. (A) and (B) show beeswarm plots of SHAP for Purity and Yield. respectively. Each point represents an individual experiment, with the position along the X-axis indicating the contribution of a parameter to the outcome. Positive values correspond to an increase in the objective, while negative values indicate a decrease. Color represents the parameter value (low to high).}
\end{figure}

After accumulating the experimental data, we plot them in the
visualizations tab. The user can plot how outcomes depend on parameters,
using 2D plots, 3D plots, and colormaps (Fig. \ref{fig:fig2}). In the explanation
tab, the user can examine how much each parameter contributes to the
objectives by plotting their SHAP (SHapley Additive exPlanations)
values.\cite{lundbergUnifiedApproachInterpreting2017} (Fig. \ref{fig:fig3}). These plots provide multi-dimensional process understanding and guide hypothesis generation. In this case study, we
show that a BASIL user optimizes their reaction in only 5 iterations, with a batch size of 10 every time.

\section{Conclusion}
BASIL is a desktop application for process optimization. It includes a modern graphical user interface and a library of surrogate models and acquisition functions for both single- and multi-objective optimization problems. It also includes the ability to visualize and explain how experimental parameters affect the process output. The software is easy to use, since no prior knowledge of Python or ML is required, making it accessible to researchers with no computer expertise. 

\section{Data and Code Availability}
Data from our example case study can be found at \url{https://github.com/lerakaneva/basil_demo_notebook_pareto}. The BASIL software is freely available under a custom UNC-Chapel Hill license. It is free for academic and non-commercial purposes, but commercial users must contact the University of North Carolina at Chapel Hill to negotiate a license agreement. The software can be downloaded at \url{https://github.com/molecularmodelinglab/BASIL}.

\section{Acknowledgments}
The development of BASIL was made possible by funding from the Food and Drug Administration (Grant R01FD007481) and the Triangle Universities Center for Advanced Studies Inc (TUCASI).

\printbibliography

% \bibliographystyle{unsrtnat}
% \bibliography{references}  %%% Uncomment this line and comment out the ``thebibliography'' section below to use the external .bib file (using bibtex) .

\end{document}